\documentclass[10pt, logo, twocolumn, copyright]{nvidiatechreport}

\usepackage{mdframed}
\usepackage[utf8]{inputenc} 
\usepackage[T1]{fontenc}    

\usepackage{amsfonts}       
\usepackage{nicefrac}       
\usepackage{microtype}      
\usepackage[dvipsnames]{xcolor}         
\usepackage[dvipsnames]{xcolor}         
\definecolor{mediumgray}{gray}{0.6}
\usepackage{multirow}
\usepackage{subcaption}
\usepackage{multirow}
\usepackage{multicol}
\usepackage{graphicx}
\usepackage[numbers]{natbib}
\usepackage{tabto}
\usepackage{xspace}
\usepackage{amsmath}
\usepackage{adjustbox}
\usepackage{enumitem}
\usepackage{wrapfig}
\usepackage{dblfloatfix}

\usepackage{footmisc}

\usepackage{float}

\usepackage{natbib}

\usepackage{hyperref}
\usepackage{url}
\usepackage{booktabs}
\usepackage{multirow}
\usepackage{enumitem}
\usepackage{adjustbox}
\usepackage{tabularx}
\usepackage{arydshln}
\usepackage{wrapfig}

\usepackage{multirow}
\usepackage{colortbl}
\usepackage{amsmath}
\usepackage{makecell}
\usepackage{hhline}
\usepackage{array}
\usepackage{diagbox}
\usepackage{graphicx}
\usepackage{adjustbox}

\usepackage{fp} 

\usepackage{xcolor}
\usepackage[subtle]{savetrees}
\usepackage{pifont}
\usepackage{caption}
\usepackage{placeins}

\DeclareRobustCommand\onedot{\futurelet\@let@token\@onedot}
\def\@onedot{\ifx\@let@token.\else.\null\fi\xspace}

\def\eg{\emph{e.g}\onedot}

\makeatother

\begin{document}

\title{Cosmos-H-Surgical: Learning Surgical Robot Policies from Videos via World Modeling}
\author{Yufan He\textsuperscript{1,$\ddag$} \qquad Pengfei Guo\textsuperscript{1,$\ddag$} \qquad Mengya Xu\textsuperscript{2,$\ddag$} \qquad Zhaoshuo Li\textsuperscript{1} \qquad Andriy Myronenko\textsuperscript{1} \qquad Dillan Imans\textsuperscript{3} \qquad Bingjie Liu\textsuperscript{4} \qquad Dongren Yang\textsuperscript{4} \qquad Mingxue Gu\textsuperscript{1} \qquad Yongnan Ji\textsuperscript{1} \qquad Yueming Jin\textsuperscript{5} \qquad Ren Zhao\textsuperscript{6} \qquad Baiyong Shen\textsuperscript{6} \qquad Daguang Xu \textsuperscript{1} \\~\\
\textsuperscript{1}NVIDIA \quad \textsuperscript{2}The Chinese University of Hong Kong \quad \textsuperscript{3}Sung Kyun Kwan University
\quad \textsuperscript{4} Wenzhou Medical University \\ \textsuperscript{5} National University of Singapore \quad \textsuperscript {6} Ruijin Hospital\\ 
\textsuperscript{$\ddag$}Equal First Author}

\correspondingauthor{X}

\begin{abstract}
Data scarcity remains a fundamental barrier to achieving fully autonomous surgical robots. While large-scale vision–language–action (VLA) models have shown impressive generalization in household and industrial manipulation by leveraging paired video-action data from diverse domains, surgical robotics suffers from the paucity of datasets that include both visual observations and accurate robot kinematics. In contrast, vast corpora of surgical videos exist, but they lack corresponding action labels, preventing direct application of imitation learning or VLA training. In this work, we aim to  alleviate this problem by learning policy models from Cosmos-H-Surgical, a world model designed for surgical physical AI. We curated the Surgical Action-Text Alignment (SATA) dataset with detailed action description specifically for surgical robots. Then we built Cosmos-H-Surgical based on the most advanced physical AI world model and SATA. It's able to generate diverse, generalizable and realistic surgery videos. We are also the first to use an inverse-dynamics model to infer pseudo-kinematics from synthetic surgical videos, producing synthetic paired video–action data. We demonstrate that a surgical VLA policy trained with these augmented data significantly outperforms models trained only on real demonstrations on a real surgical robot platform. Our approach offers a scalable path toward autonomous surgical skill acquisition by leveraging the abundance of unlabeled surgical video and generative world modeling, thus opening the door to generalizable and data-efficient surgical robot policies.

\smallskip
\textbf{Code:}
\href{https://github.com/NVIDIA-Medtech/Cosmos-H-Surgical}{https://github.com/NVIDIA-Medtech/Cosmos-H-Surgical}

\textbf{Model:}
\href{https://huggingface.co/nvidia/Cosmos-H-Surgical}{https://huggingface.co/nvidia/Cosmos-H-Surgical}
\end{abstract}
\twocolumn[{%
\renewcommand\twocolumn[1][]{#1}%
\maketitle
\begin{center}
    \captionsetup{type=figure}
    \includegraphics[width=1\textwidth]{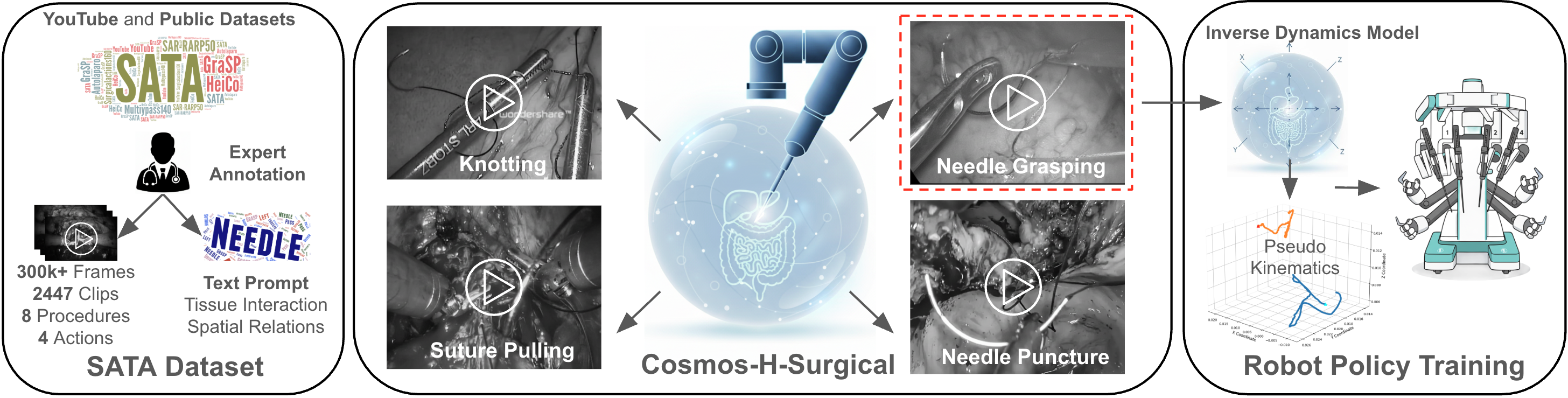}
    \captionof{figure}{We curate SATA dataset with surgical videos and detailed text annotations for physical AI. A powerful world model (Cosmos-H-Surgical) is built using Cosmos2.5~\cite{nvidia2025worldsimulationvideofoundation} and SATA, which is able to generate high quality, generalizable videos for surgical robots. We are also the first to illustrate the efficacy of surgical world modeling for autonomous surgical robots.}
\end{center}%
}]

\section{Introduction}
\label{sec:intro}
\begin{figure*}
    \centering
    \includegraphics[width=1\linewidth]{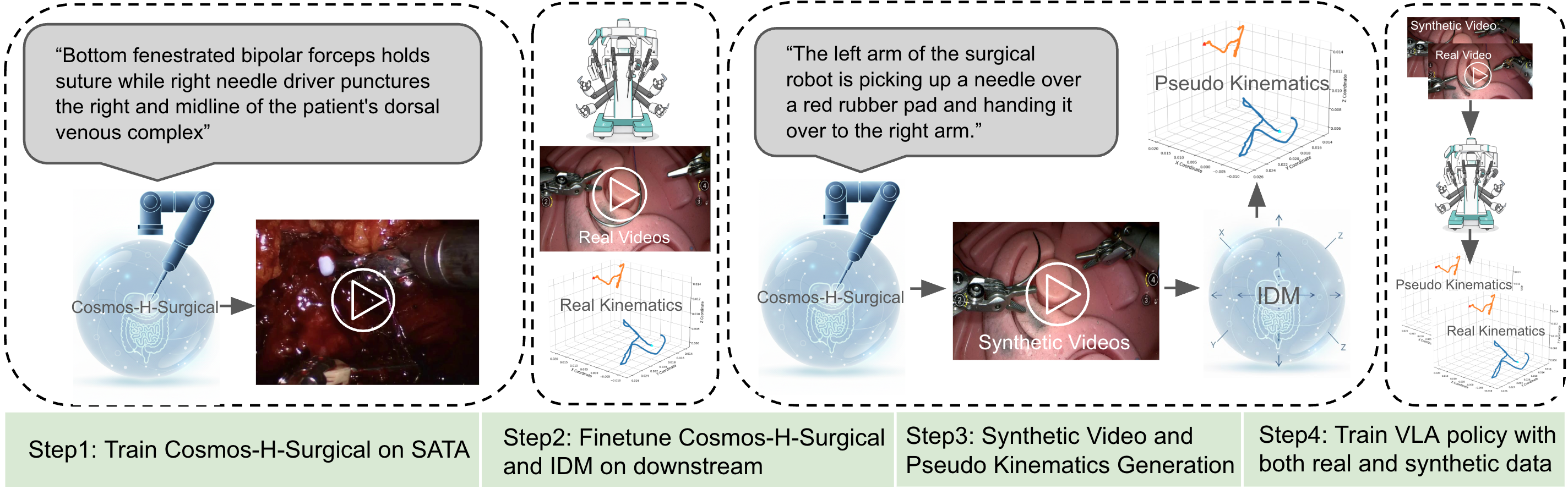}
    \caption{The overall workflow. The Cosmos-H-Surgical model is first pretrained with large scale surgical videos with text annotations, based on Cosmos 2.5~\cite{nvidia2025worldsimulationvideofoundation}. For downstream task with specific robot type and task, we finetune Cosmos-H-Surgical and train the inverse dynamic model~(IDM) for the specific embodiment. In step 3 we generate synthetic video rollouts from Cosmos-H-Surgical and get pseudo kinematics from the IDM. We use both real data and synthetic data to train the surgical VLA model.}
    \label{fig:workflow}
\end{figure*}

Autonomous robotic surgery promises to enhance precision, reduce surgeon fatigue, and scale complex procedures. Yet, despite extensive research, training robust robotic policies for surgical manipulation remains exceptionally challenging. A major bottleneck is the lack of large, diverse datasets that include both high-fidelity visual observations (e.g., endoscopic video) and synchronized robot kinematics or control commands. Collecting such paired demonstrations is prohibitively expensive, constrained by operating room access, patient safety, and regulatory hurdles.

In parallel, the robotics community has seen tremendous progress in large vision–language–action (VLA) models. Models such as RT-2~\cite{brohan2023rt2}, OpenVLA~\cite{kim2024openvla}, and GR00T~\cite{bjorck2025gr00t} has demonstrated the potential of foundation models to generalize across diverse robotic manipulation tasks. These models leverage large-scale multimodal datasets that couple visual observations, language descriptions, and robot actions, enabling rich world understanding and robust policy learning, yielding significantly improved generalization to novel objects and commands in non-surgical manipulation domains. Similarly, recent work on synthetic data generation such as DreamGen~\cite{jang2025dreamgen} shows that video world-models combined with inverse dynamics can produce synthetic paired video–action datasets that boost policy learning in general manipulation tasks.

However, extending such capabilities to surgical robotics remains a significant challenge. Unlike household or industrial domains, surgical robotics suffers from severe data scarcity, particularly the lack of large-scale datasets with synchronized visual and kinematic information. Surgical data collection is constrained by privacy regulations, ethical considerations, and the cost of robotic surgical systems, limiting the ability to train data-hungry models such as VLAs and imitation learning (IL) policies. Synthetic physics-based simulators~\cite{schmidgall2024surgical,xu2021surrol} attempt to fill the gap, but often suffer from a large visual and dynamic domain shift to real surgical systems and lacking soft body simulation, limiting policy transfer.

To address these limitations, we propose \textbf{Cosmos-H-Surgical}, a unified framework that leverages a surgical world model to enable scalable policy learning. Specifically, we curate and annotate the Surgical Action–Text Alignment (SATA) dataset with expert-labeled surgical video clips covering four core surgery actions with detailed spatial, interaction, and anatomical context. Using SATA, we train a diffusion-based world model~\cite{nvidia2025worldsimulationvideofoundation} capable of generating photorealistic, task-consistent surgical scenes. We then employ an inverse dynamics model (IDM) to infer pseudo-kinematics, producing synthetic paired video–action data that can be directly used for surgical VLA policy learning. We validate our approach on a surgical robotic platform performing needle pick-up and hand-over tasks—a fundamental and dexterous actions representative of real surgical manipulation. We use both real surgical demonstrations with kinematic supervision and synthetic videos labeled with pseudo-actions to train the GR00T N1.5 VLA model~\cite{bjorck2025gr00t}. Our results demonstrate that incorporating synthetic world model data leads to substantial improvements in policy performance with lower trajectory prediction error.

While prior works such as DreamGen~\cite{jang2025dreamgen} explored world-model-based learning from synthetic videos, these domains lack the unique visual and physical complexity of surgery, where specular tissue surfaces, endoscopic occlusion, and constrained tool motion present distinct modeling challenges.
Recent efforts have begun addressing this gap within surgical contexts: GAS~\cite{lin2024world} applies world-model-based reinforcement learning for grasping, SurgWM~\cite{koju2025surgical} generates controllable surgical videos with dynamic prediction, and the Suturing World Model~\cite{turkcan2025towards} forecasts tool trajectories for automated suturing.
However, these approaches remain limited to narrow tasks or visual prediction without explicit integration of text grounding and kinematics. 
Cosmos-H-Surgical is the \emph{first} to integrate large-scale text-aligned surgical video modeling with pseudo-kinematics generation for embodied policy learning and bridging the gap between unlabeled surgical videos and robot actions.
Scalable data generation without collecting in-vivo trajectories can dramatically accelerate surgical autonomy while maintaining patient safety. Our framework addresses the core bottleneck of data scarcity in surgical robotics and opens a scalable path toward autonomous surgical skill acquisition.

In summary, our key contributions are:
\begin{enumerate}
\item We curate the Surgical Action–Text Alignment (SATA) dataset, a large-scale surgical video–text corpus comprising 2,447 expert-annotated video clips (over 300k frames) that capture fine-grained spatial relationships and tool–tissue interactions across 8 procedures, designed specifically to support the development of physical AI models. 
\item We develop the first surgical world model that's based upon state-of-the-art physical AI world models, and finetuned with SATA, demonstrating strong generalizability, high video quality, and realistic dynamics.  
\item We are the first to connect surgical world models with robot learning by synthesizing video–action data using inverse dynamics models, achieving substantial performance improvements in surgical robot learning.
\end{enumerate}
Our results highlight the potential of generative world modeling to complement real surgical data and pave the way for foundation models that enable scalable, autonomous, and safe surgical policy learning.

\section{Related Work}

\noindent\textbf{VLA Models and Imitation Learning~(IL).}
Large-scale vision–language–action (VLA) models have recently emerged as a powerful paradigm for general-purpose robotic policy learning. Recent open-source efforts such as OpenVLA~\cite{kim2024openvla}, $\pi_0$~\cite{black2410pi0}, and GR00T N1~\cite{bjorck2025gr00t} have been trained on massive and diverse data sources, leveraging diverse robot embodiments and multi-task datasets and can robustly handle real-world variability. To train those VLAs, imitation learning (IL) remains the most direct and scalable approach, typically via behavior cloning (BC)~\cite{pomerleau1989alvinn}. However, BC suffers from covariate shift, producing rapidly worsening performance when the dataset is limited~\cite{ross2011dagger}. In robotics, several works aim to improve data efficiency using offline datasets~\cite{kumar2020offline} or synthetic augmentation~\cite{james2019sim2real}.
However, despite these advances, current VLA models still rely heavily on imitation learning on large scale paired image–action datasets, which are scarce in specialized domains such as surgical robotics.

\noindent\textbf{Surgical World Models and Video generation.}
Learning compact models of the world has long been a goal of model-based reinforcement learning (MBRL).
World Models~\cite{ha2018worldmodels} demonstrated that compact latent dynamics models can simulate plausible trajectories for policy learning.
Subsequent works such as PlaNet~\cite{hafner2019planet}, Dreamer~\cite{hafner2020dreamer}, and DreamerV3~\cite{hafner2023dreamerv3} extended these ideas to high-dimensional visual environments, enabling agents to imagine future rollouts and train in latent space.
In the surgical domain, video generation and world modeling are only beginning to emerge.
Endora~\cite{li2024endora} integrates a spatiotemporal transformer with a latent diffusion backbone for endoscopic video synthesis.
SurGen~\cite{cho2024surgen} proposes a text-guided diffusion model tailored to laparoscopic cholecystectomy.
VISAGE~\cite{yeganeh2024visage} formulates future surgical video prediction conditioned on a single frame and an action scene graph, modeling tool–tissue interactions for temporally coherent generation. GAS~\cite{lin2024world} applies world-model-based reinforcement learning for robust grasping across diverse objects.
SurgWM~\cite{koju2025surgical} introduces controllable surgical video generation and interactive dynamics prediction from visual inputs alone.
Suturing World Model~\cite{turkcan2025towards} learns task-specific dynamics to anticipate tool trajectories during automated suturing. Cosmos-Surg-dVRK~\cite{zbinden2025cosmos} focuses on policy evaluation using action conditioned video generation model.
While these studies mark important progress, they are limited to single-task or object-specific scenarios, and rely on narrowly scoped datasets lacking high-quality text–action alignment or procedural diversity, and many are not open-sourced, limiting reproducibility and broader impact.
In contrast, our approach leverages the curated SATA dataset to train a surgical world model capable of generating photorealistic, task-consistent videos explicitly designed for physical AI and downstream robot policy learning.

\noindent\textbf{Learning Policy Models from Videos.}
Learning policy models directly from video has become increasingly important due to the lack of datasets with kinematics.  Ye~\cite{ye2024latent} and Jang~\cite{jang2025dreamgen} propose a latent-action pretraining scheme using internet-scale videos without robot action labels to bootstrap vision-language-action models. 
Hu~\cite{hu2025videoprediction} train predictive visual representations via video diffusion models and embed an implicit inverse-dynamics policy conditioned on these representations. 
Bharadhwaj~\cite{bharadhwaj2025gen2act} use human-video generation to condition robot policies on novel scenarios, reducing reliance on robot-collected data. 
Li~\cite{li2025unifiedvideoaction} present a unified video-action latent model that jointly handles forward/inverse dynamics, video generation, and action inference within one framework. 
Tian~\cite{tian2025predictiveinverse} close the loop by using inverse-dynamics models conditioned on predicted visual states for large-scale manipulation training. These developments directly inspire our approach, which uses IDM to connect world models with surgical robot learning.

\noindent\textbf{Automated Surgical Robotics.}
Automation in surgical robotics have increasingly leveraged learning-based approaches. Long~\cite{long2025surgical} introduced a vision-based embodied intelligence framework that enables zero-shot sim-to-real transfer for a variety of laparoscopic assistive tasks using imitation learning, Kim~\cite{kim2024surgical} proposed the Surgical Robot Transformer (SRT), which addresses the da Vinci robot’s inaccurate kinematics by formulating actions in a hybrid-relative space, achieving high success rates on fundamental tasks. The same group later developed SRT-H~\cite{kim2025srt} that enables step-level autonomy in complex procedures such as cholecystectomy. While these works demonstrate impressive progress in task autonomy, they also reveal a critical need for large-scale, diverse surgical demonstration datasets, which in surgery is uniquely challenging: obtaining large-scale paired visual–kinematic datasets requires specialized hardware and surgeon supervision.

\section{Method}
\label{sec:method}
The overall workflow is shown in Fig.~\ref{fig:workflow}. The surgical world model is obtained by finetuning Cosmos-Predict2.5~\cite{nvidia2025worldsimulationvideofoundation} model on surgical videos with detailed annotations. Then we demonstrate its efficacy in downstream surgical robot tasks. Since the world model has not seen specific surgical robot embodiments, we finetune the world model on robots and task specific data. Meanwhile we build an inverse dynamic model~(IDM) for this specific robot. The world model generates video rollouts and IDM model label those videos with pseudo action kinematics.

\subsection{Dataset Curation}\label{sec:dataset_curation}

\noindent\textbf{Surgical Action–Text Alignment (SATA) Dataset.}  
We introduce \textbf{SATA}, a large-scale surgical action–text alignment dataset comprising 2,447 expert-annotated video clips (over 300k frames) collected across 8 different surgery types.
Each clip captures one of four fundamental actions: needle grasping (689), needle puncture (989), suture pulling (475), and knotting (294), which provides diverse visual and procedural coverage. SATA is curated by aggregating and re-annotating videos from credentialed YouTube surgical channels~\cite{schmidgall2024general} and publicly available datasets, including GraSP~\cite{ayobi2024pixel}, SAR-RARP50~\cite{psychogyios2023sar}, Multiypass140~\cite{ramesh2023weakly}, SurgicalActions160~\cite{schoeffmann2018video}, AutoLaparo~\cite{wang2022autolaparo}, and HeiCo~\cite{maier2021heidelberg}.  
Unlike surgical VLM datasets, such as SurgVLM-DB~\cite{zeng2025surgvlm}, which focus primarily on semantic reasoning and instruction following, SATA is specifically designed for \textit{physical AI}: its fine-grained action labels and detailed text descriptions capture precise tool–tissue interactions and spatial relationships needed for training world models.

The four action categories are defined by decomposing the suturing procedure into its fundamental steps and annotated according to the following criteria:  
\begin{itemize}[leftmargin=*]
    \item \textbf{Needle grasping:} Approaching and securing the needle with a smooth, controlled trajectory, emphasizing the dynamic ``go-to-grasp'' motion rather than the subsequent static hold.
    \item \textbf{Needle puncture:} Inserting the needle into tissue with precise control over its entry angle and depth.
    \item \textbf{Suture pulling:} Drawing the suture thread through tissue after puncture completion, typically by pulling on the needle or thread.
    \item \textbf{Knotting:} Looping and tightening the suture material to secure the tissue layers together.
\end{itemize}

Each clip is paired with a rich textual description detailing (i) spatial relationships between surgical instruments, (ii) the anatomical structure being manipulated, and (iii) the description of instrument–tissue interaction.  
For example: “The left needle driver punctures the right side of the patient’s dorsal venous complex.” Additional statistics and dataset breakdowns for SATA are provided in the supplementary materials.

\noindent\textbf{Real World Trajectories.} For real-world validation, we aim to demonstrate that synthetic videos generated by the world model can enhance autonomous surgical robot policy learning. Due to the high cost and regulatory constraints of in-vivo experiments, we evaluate our method using the “Needle Pickup and Hand-Over” task on rubber pad. The experiments are conducted on a commercial endoscopic surgical system (robot and manufacturer anonymized), which consists of a stereo endoscope and two articulated robotic forceps (left and right arms). During the task, the left arm grasps the needle tip and hand it over to the right arm.

We collected a total of 60 successful human-teleoperated demonstrations for training and testing. Each episode includes synchronized endoscopic video (average length: 217 frames) and corresponding action kinematics. In addition to these task-specific demonstrations, we also utilized 66 out-of-domain episodes (around 60k action frames pairs) depicting general robot movements unrelated to the needle pickup task. These data are used to pretrain a foundational inverse dynamics model (IDM) for the surgical robot, providing transferable motion understanding across tasks.

\begin{figure}[t!]
    \centering
    \includegraphics[width=0.95\linewidth]{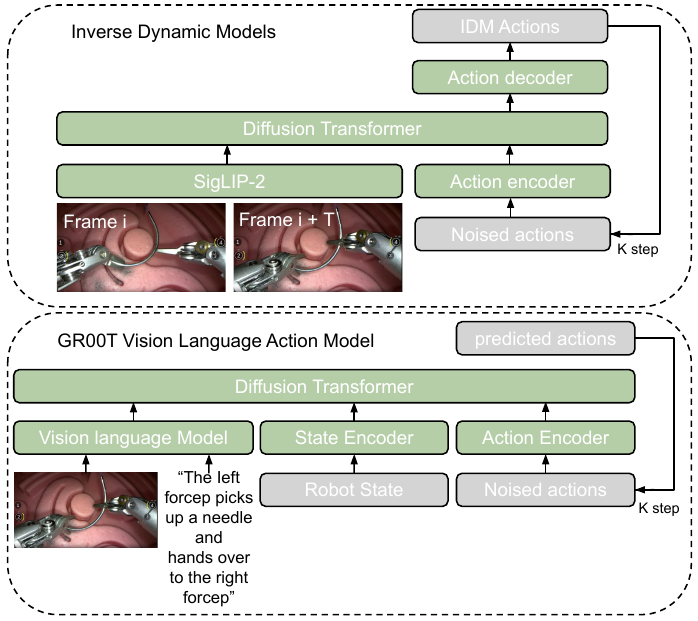}
    \caption{The model architecture for inverse dynamic models~(IDM) and the vision language action foundation model (GR00T N1.5). They share similar architectures but IDM does not use text prompt nor robot state.}
    \label{fig:models}
\end{figure}

The robot states are the same as the action kinematics with the same dimension and values.  The action kinematics at each timestep is represented as a 20-dimensional continuous vector:
\[
\mathbf{a}_t = [\mathbf{p}_L, \mathbf{r}_L, g_L, \mathbf{p}_R, \mathbf{r}_R, g_R],
\]
where each term encodes the motion of the left (L) and right (R) instruments relative to the endoscope frame.

Specifically,
\begin{itemize}
    \item $\mathbf{p}_L = [x_L, y_L, z_L] \in \mathbb{R}^3$ represents the translational offset (cartesian coordinates) of the left forcep tip with respect to the endoscope frame (represented by meters).
    \item $\mathbf{r}_L = [r_{L1}, \dots, r_{L6}] \in \mathbb{R}^6$ denotes the 6D rotation representation of the left instrument’s end-effector orientation. The 6D rotation formulation~\cite{zhou2019continuity} is used to avoid discontinuities and ensure smooth interpolation in $\mathrm{SO}(3)$ by dropping the last column of the rotation matrix. 
    \item $g_L \in \mathbb{R}$ indicates the gripper jaw opening in radians.
\end{itemize}

The same definitions apply for the right forcep with $\mathbf{p}_R$, $\mathbf{r}_R$, and $g_R$. 
This yields a total of 20 control dimensions. 
All translation and rotation components are expressed relative to the endoscope’s coordinate frame to ensure view-consistent control.

\subsection{Surgical World Model}\label{sec:worldmodel}

In this work, we adopt Cosmos-Predict2.5~\cite{nvidia2025worldsimulationvideofoundation}, a large-scale video world model pretrained on diverse robotic and embodied datasets, as our base model and adapt it to the surgical domain. Cosmos-Predict2.5~\cite{nvidia2025worldsimulationvideofoundation} leverages diffusion-based latent video prediction with transformer backbones to simulate high-fidelity spatiotemporal dynamics. Its large-scale pretraining on heterogeneous robotic and human teleoperation videos provides strong priors for object interactions, tool motion, and scene dynamics, making it particularly well-suited for domains with limited labeled data, such as surgical robotics. By leveraging these pretrained representations, we can efficiently transfer general video dynamics knowledge to the endoscopic surgical domain, reducing the amount of domain-specific data required while preserving temporal coherence and realism. To adapt the pretrained model to the surgical domain, we fine-tune it on the curated SATA dataset and real-world surgical trajectories described in previous section, which enables the model to capture the unique visual dynamics of robotic endoscopic scenes, such as instrument–tissue interaction, limited field-of-view motion, and constrained articulation patterns. 
Unlike policy models, our model conditions only on the first observed frame $I_0$ and predicts future trajectories, capturing the temporal evolution of the surgical scene.

We adopt Low-Rank Adaptation (LoRA)~\cite{hu2022lora} to efficiently specialize Cosmos-Predict2.5~\cite{nvidia2025worldsimulationvideofoundation} for the surgical domain while preserving its general video modeling capabilities. LoRA modules are inserted into the transformer’s attention and feed-forward layers, enabling parameter-efficient finetuning with minimal forgetting.
During adaptation, the model learns to predict future latent video frames conditioned on an initial observation and text prompt.
Given an initial frame $I_0$, the world model produces a rollout $\hat{I}{1:T} = \mathcal{W}{\theta}(I_0)$, where $\mathcal{W}_{\theta}$ denotes Cosmos-Predict2.5~\cite{nvidia2025worldsimulationvideofoundation} augmented with LoRA adapters.
A spatiotemporal encoder extracts features from $I_0$, a transformer-based latent dynamics module models temporal evolution, and a decoder reconstructs the predicted frames. We adopt the Flow Matching (FM) formulation~\cite{lipman2022flow} to train the surgical video world model due to its conceptual simplicity and practical effectiveness. More training detail of Cosmos-H-Surgical can be found in the supplementary material.

\subsection{Inverse Dynamic Models and Policy Models}

We follow the DreamGen IDM design~\cite{jang2025dreamgen,baker2022video} and use GR00T N1.5~\cite{bjorck2025gr00t} as the policy model. The model architectures are shown in Fig.~\ref{fig:models}. These two models predict robotic actions with DIT~\cite{peebles2023scalable} and flow matching heads~\cite{lipman2022flow}.  The major difference is that IDM's inputs are two frames from the same video (T = 16 frames apart), and the model predicts the robot actions for every frame between these two input frames, while the GR00T policy model takes the current frame and text prompt, together with the robot state to predict the actions for future 16 frames.

\section{Experiments}\label{experiments}
\subsection{Surgical World Model Evaluation}\label{sec:worldmodel_eval}
We first evaluate the proposed \textit{Comos-H-Surgical} world model on (i) video generation quality using the curated \textbf{SATA} dataset of internet surgical videos and (ii) few-shot adaptation to collected real trajectories. The goal is to assess both perceptual fidelity and transferability to downstream policy learning.

\begin{figure}[t!]
    \centering
    \includegraphics[width=0.95\columnwidth]{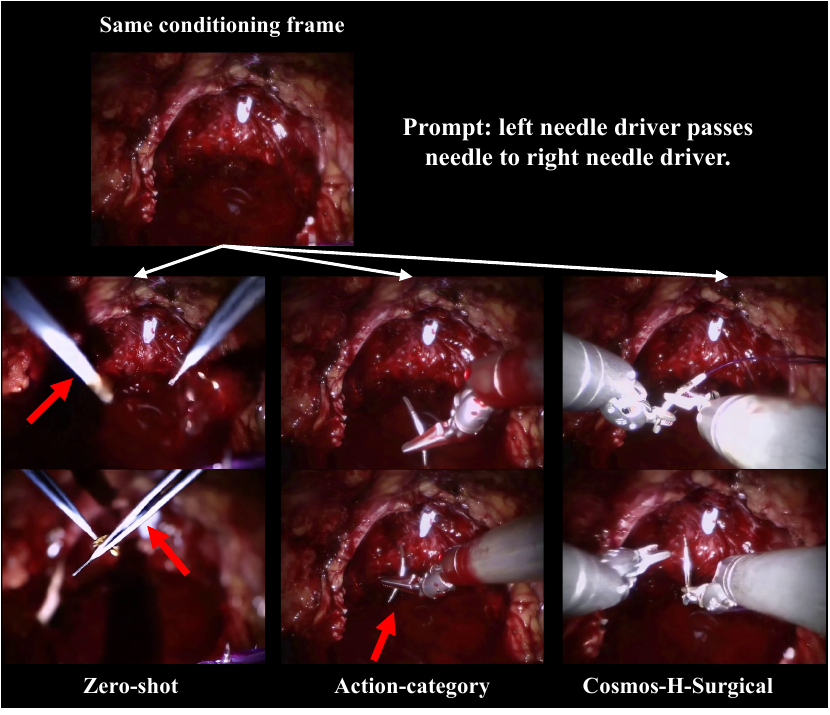}
    \caption{
    \textbf{Qualitative comparison of three variants of Cosmos-Predict2.5~\cite{nvidia2025worldsimulationvideofoundation} on the SATA dataset.} Red arrows highlight incorrect surgical tools or actions in the generated frames.
    }
    \label{fig:vis_compare}
\end{figure}

\begin{figure*}[htbp!]
    \centering
    \includegraphics[width=1\textwidth]{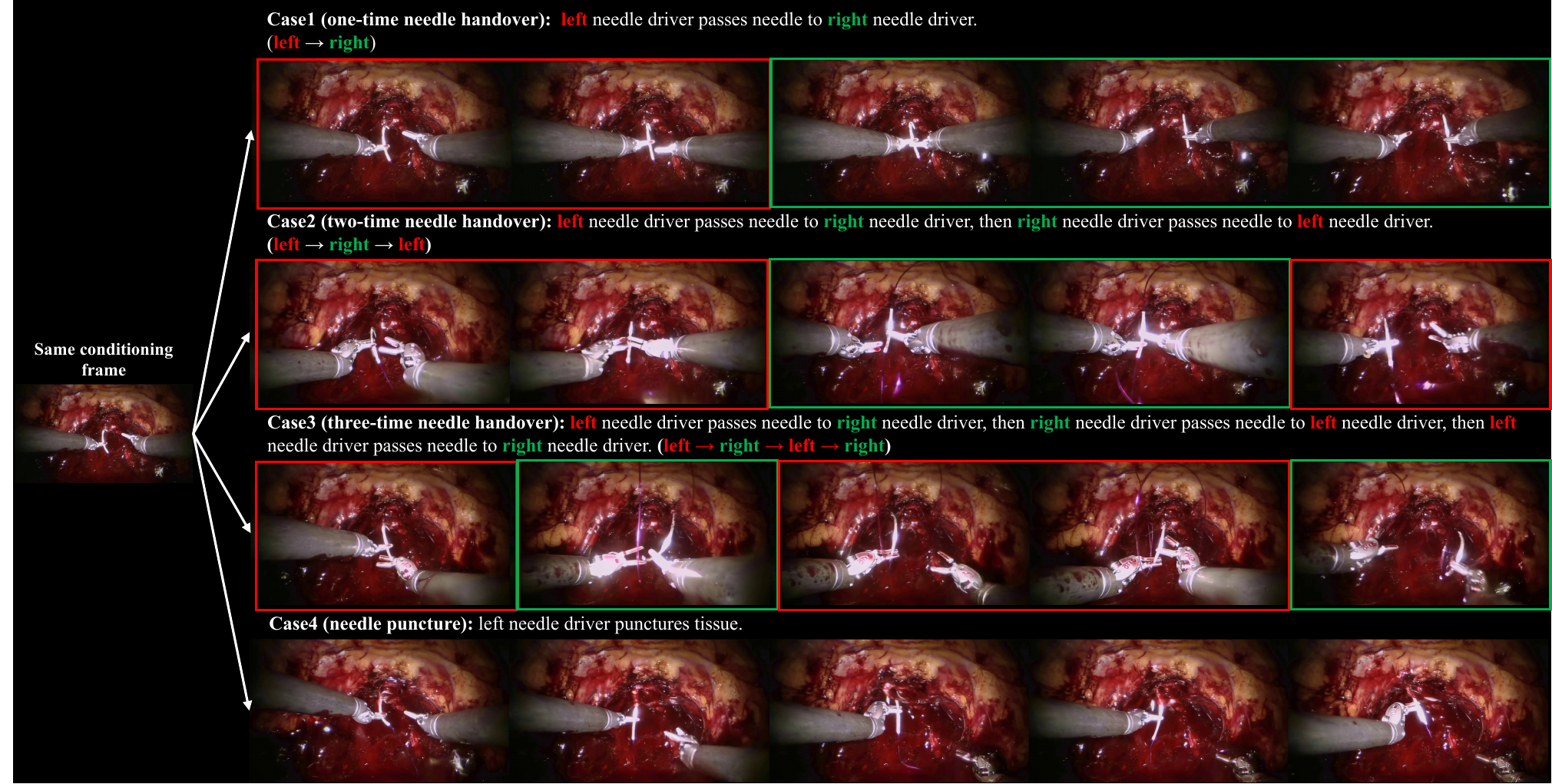}
    \caption{
    \textbf{New behavior generalization via strong text–video alignment.} Given the same conditioning frame, our surgical world model generates distinct video rollouts corresponding to four task prompts: (1) one-time needle handover, (2) two-time needle handover, (3) three-time needle handover, and (4) needle puncture. 
    }
    \label{fig:text_video_aligment_example}
\end{figure*}

\noindent\textbf{Video Generation on SATA.} We evaluate three variants of Cosmos-Predict2.5~\cite{nvidia2025worldsimulationvideofoundation}:
(1) \textbf{Zero-Shot}, using the base model evaluated without any domain adaptation;
(2) \textbf{Action-Category}, finetuned using coarse, category-level captions where all videos within the same action category share an identical prompt; and
(3) \textbf{Cosmos-H-Surgical}, finetuned on SATA’s fine-grained, expert-curated textual descriptions capturing detailed tool–tissue interactions and spatial relations.
We report Fréchet Video Distance (FVD)~\cite{unterthiner2019fvd} and follow~\cite{nvidia2025worldsimulationvideofoundation} to report the three representative VBench~\cite{huang2024vbench} metrics. 
As shown in Table~\ref{tab:bsa_quantitative}, Cosmos-H-Surgical trained with curated prompts achieves the lowest FVD and highest alignment scores, indicating substantial gains in perceptual realism and semantic coherence over zero-shot and coarse category-level prompt baselines. Qualitative results in Figure~\ref{fig:vis_compare} further highlight these differences in a challenging scenario where the initial frame contains no visible surgical tools.
The Zero-Shot model hallucinates an incorrect instrument due to limited domain priors, and the Action-Category model initiates a wrong action (tissue puncture).
In contrast, the Cosmos-H-Surgical correctly follows the textual instruction and completes the intended needle grasping motion with consistent behavior.

\begin{table}[!t]
\centering
\small
\caption{\textbf{Quantitative evaluation of surgical video generation on the curated SATA dataset.} We report FVD and VBench metrics: dynamic degree (DD), imaging quality (IQ), and overall consistency (OC).
}
\begin{tabular}{c|cccc}
\hline
Method          & FVD $\downarrow$ & DD $\uparrow$ & IQ $\uparrow$ & OC $\uparrow$ \\ \hline
Zero-shot       & 175.4            & 26.9          & 48.7          & 18.0          \\
Action-category & 143.0            & 26.5          & 49.0          & 18.1          \\
Cosmos-H-Surgical       & \textbf{106.5}   & \textbf{62.4} & \textbf{49.3} & \textbf{21.5} \\ \hline
\end{tabular}
\label{tab:bsa_quantitative}
\end{table}

\noindent\textbf{New Behavior Generalization.} To evaluate the model’s capacity to generalize to diverse and semantically consistent actions, we test its response to varying textual prompts describing distinct surgical behaviors. Figure~\ref{fig:text_video_aligment_example} presents representative video rollouts generated from the same conditioning frame under four distinct textual prompts: one-time/two-time/three-time needle handover, and needle puncture. 
The multi-step handover cases are particularly noteworthy, because during data curation all multi-time handovers are decomposed into single handover segments. Therefore, the two- and three-time handover sequences represent \textit{novel compositions} not explicitly observed during training.
The model accurately follows each instruction, producing coherent sequences that reflect increasing motion complexity while preserving visual realism. These results demonstrate strong text–video alignment and show that the Cosmos-H-Surgical can recombine learned primitives to generate anatomically plausible and temporally consistent surgical behaviors from prompt-level conditioning.

\begin{figure}[ht!]
    \centering
    \includegraphics[width=0.95\linewidth]{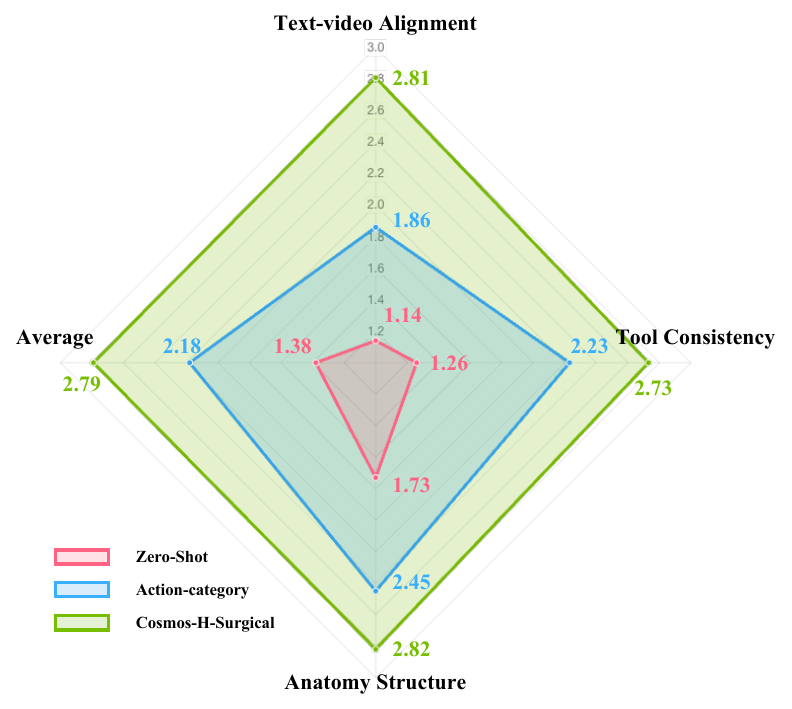}
    \caption{
    \textbf{Human expert evaluation of generated surgical videos.} Radar plot summarizing expert ratings across three criteria using a 1–3 quality scale. 
    }
    \label{fig:human_eval}
\end{figure}
\noindent\textbf{Human Expert Evaluation.}
While quantitative metrics such as FVD and VBench capture perceptual and temporal quality, they fail to fully reflect the \textit{clinical realism} required in surgical video generation. To bridge this gap, we conducted a human evaluation study to assess the anatomical plausibility, instrument behavior, and semantic faithfulness of generated videos from a surgical perspective. Three surgical experts (one surgeon supervising two residents) independently evaluated 50 video samples generated by different world model variants using a \textbf{three-level clinical quality rubric} (scores from 1 to 3, higher is better) across the following criteria:

\begin{itemize}[leftmargin=*]
    \item \textbf{Text–Video Alignment:} Assesses whether the generated scenes match the textual prompt. 

    \item \textbf{Tool Consistency:} Evaluates whether surgical instruments correspond to the prompt and remain physically consistent throughout the video sequence.  

    \item \textbf{Anatomical Structure:} Measures the plausibility of tissue and organ appearance and reactions to interaction.  
\end{itemize}
A detailed description of the rating rubric for each score level is in the supplementary materials.
Figure~\ref{fig:human_eval} summarizes the averaged scores using a radar plot. The Cosmos-H-Surgical achieves the highest ratings across all dimensions, reflecting superior text grounding, consistent instrument manipulation, and anatomically realistic tissue behavior. In contrast, the Zero-shot and Action-category variants exhibit weaker temporal coherence and less stable tool–tissue interaction, underscoring the importance of prompt-level curation for clinically faithful video generation.

\begin{table}[!t]
\centering
\small
\caption{\textbf{Few-shot finetuning results on real surgical trajectories.} We report task success rate (SR) and image quality metrics: FVD and VBench metrics. \textbf{FT} and \textbf{PT} indicate whether the model is finetuned on the 5 real trajectories or pretrained on the SATA dataset, respectively.
}
\resizebox{\columnwidth}{!}{
\begin{tabular}{ccc|ccccc}
\hline
Method         & FT & PT & SR $\uparrow$ & FVD $\downarrow$ & DD $\uparrow$ & IQ $\uparrow$ & OC $\uparrow$ \\ \hline
Zero-shot      & \ding{55}   & \ding{55}   & 0.0           & 235.2            & 53.6          & 70.3          & 20.1          \\
Finetuned-Orig & \ding{51}  & \ding{55}    & 51.8          & 212.5            & 85.7          & 72.0          & 21.1          \\
Cosmos-H-Surgical      & \ding{51}   & \ding{51}    & \textbf{73.2} & \textbf{207.1}   & \textbf{89.3} & \textbf{73.3} & \textbf{22.4} \\ \hline
\end{tabular}
 }
\label{tab:fewshot}
\end{table}

\noindent\textbf{Few-Shot Adaptation.} We further evaluate the few-shot finetuning capability of the proposed surgical world model using only \textbf{5 real-world trajectories} from the \textit{Needle Pick-Up and Hand-Over} task (Section~\ref{sec:dataset_curation}).
Three configurations are compared: (1) \textbf{Zero-Shot}, using the original model directly; (2) \textbf{Finetuned-Orig}, where the model is finetuned from the original  Cosmos-Predict2.5~\cite{nvidia2025worldsimulationvideofoundation} checkpoint; and (3) \textbf{Cosmos-H-Surgical}, where the model is first pretrained on the curated SATA dataset and then finetuned on the 5 trajectories. For evaluation, we generate videos from \textbf{56 hold-out initial frames} selected from the 66 out-of-domain episodes described in Section~\ref{sec:dataset_curation}. These frames are chosen based on proper initial needle and forceps configurations to ensure physically meaningful task initialization and comparable scene context.
We report success rate (SR) and the same suite of video quality metrics as in Table~\ref{tab:bsa_quantitative}.
To assess the success rate, surgical experts evaluate the completeness of the trajectories in the generated videos.


As shown in Table~\ref{tab:fewshot}, Cosmos-H-Surgical achieves the best overall performance, with a 73.2\% success rate and the lowest FVD among all variants. Compared to direct finetuning from the original Cosmos-2.5 checkpoint, SATA pretraining yields consistently better video quality metrics, indicating improved temporal stability and perceptual fidelity. These results demonstrate that large-scale surgical video pretraining substantially enhances the model’s ability to adapt from limited real-world data, enabling robust and data-efficient generation for downstream surgical policy learning.

\begin{figure}
    \centering
    \includegraphics[width=0.75\linewidth]{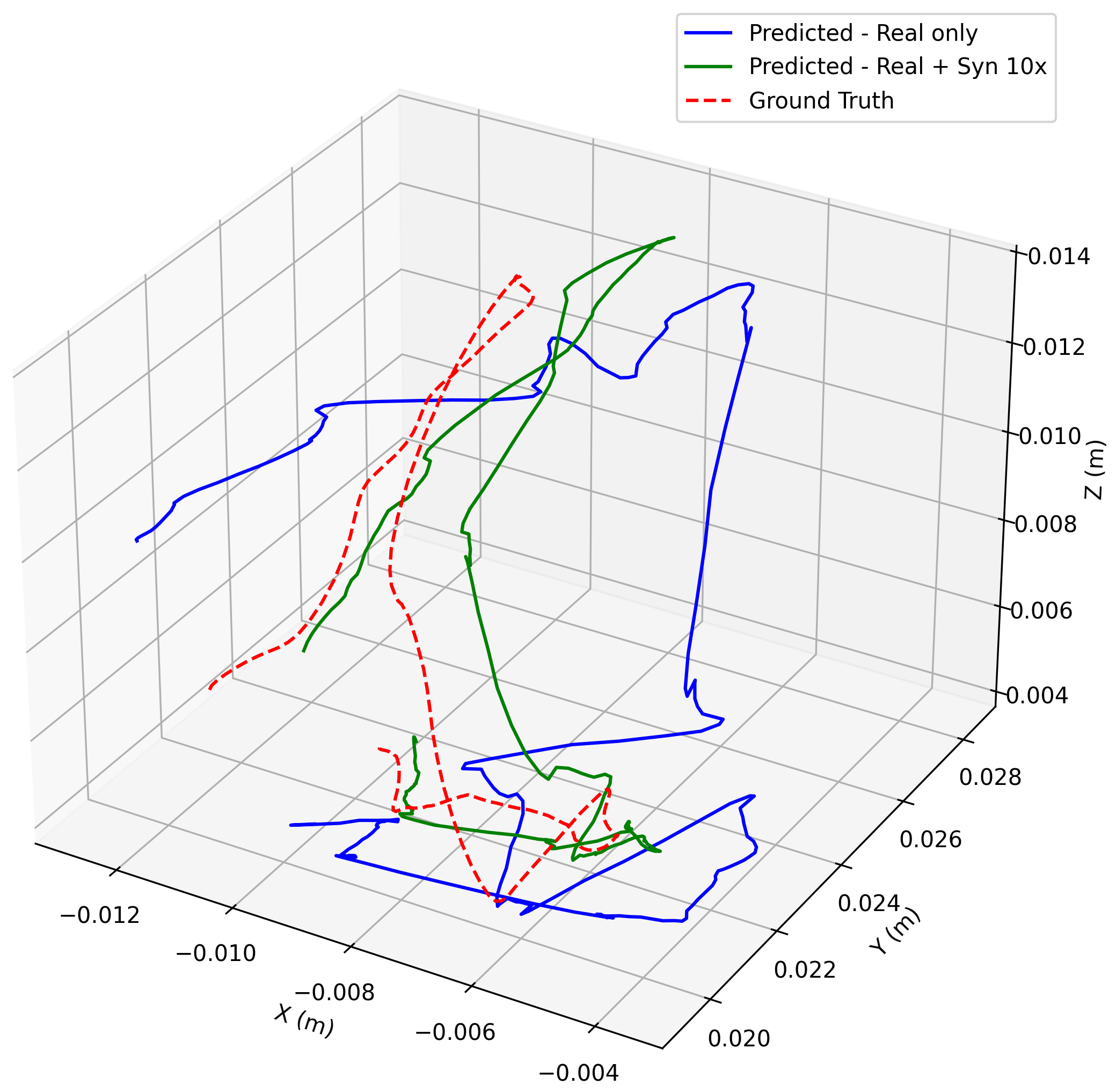}
    \caption{An example of left arm cartesian trajectory (first 3-dim of action space). Comparing Real only (Blue), Real + Syn 10x~(Green), and groundtruth~(Red).}
    \label{fig:trajectory}
\end{figure}

\begin{figure*}[ht!]
    \centering
    \includegraphics[width=1\linewidth]{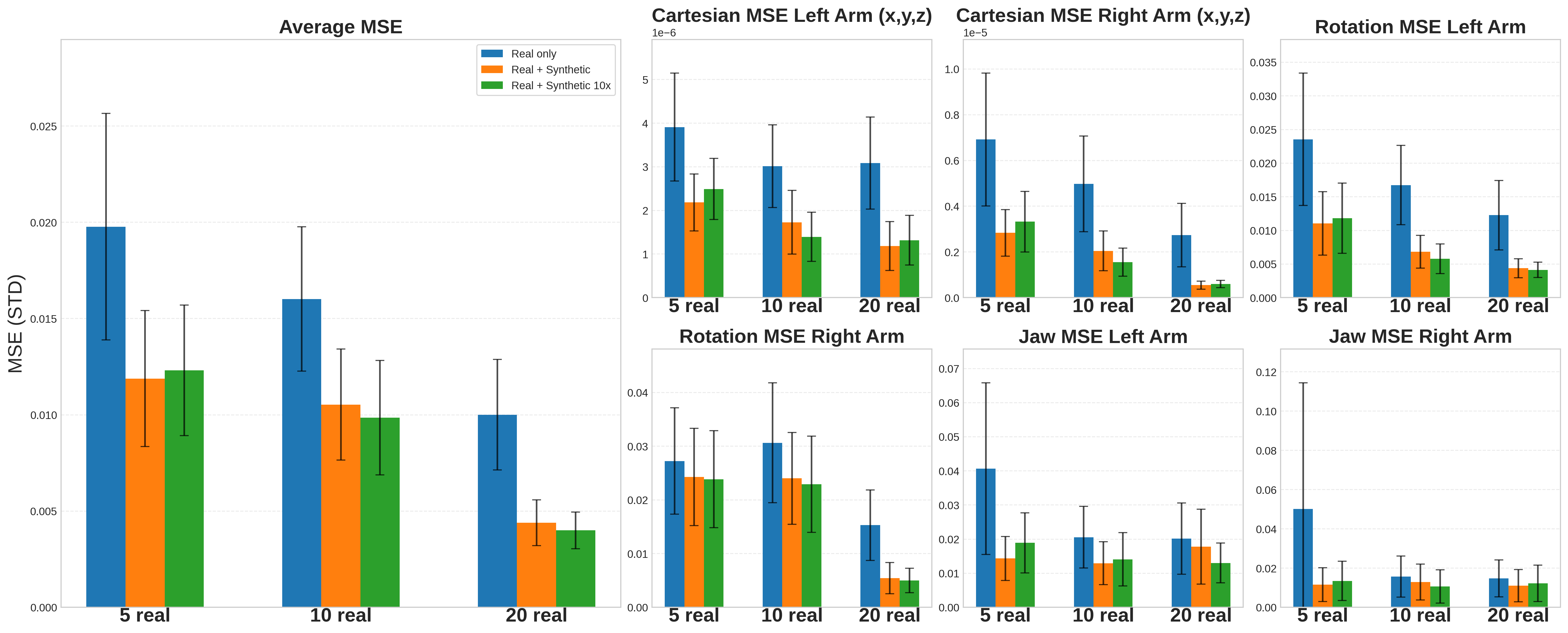} 
    \caption{Trajectory MSE (Standard deviation) on 40 test data. We use 5, 10, 20 real data to finetune the policy, starting from GR00TN1.5 pretrained checkpoint (Real), and checkpoints pretrained from 56 (Real + Synthetic) and 560~(Real + Synthetic 10x) synthetic data.}
    \label{fig:mse_comparison}
\end{figure*}

\subsection{Robotic Policy Experiments}
For the 60 human teleoperated data from surgical robots, we split the last 40 as held out test set. For the rest of the data we perform experiments by gradually increasing the training data number. We use 5, 10, and 20 data for finetuning the world model, IDM, and the GR00T policy. All videos are resized to 224$\times$224 and augmented with color jitter, and the jaw opening and cartesian actions are normalized with min-max normalization. 

\noindent\textbf{World Model Rollouts. } We finetune Cosmos-H-Surgical using 5, 10, and 20 real-world videos and each finetuned model is used to generate synthetic videos conditioned on the initial frames from the out-of-domain episodes. In total, two sets of \textbf{56 single-rollout} (1$\times$) and \textbf{560 multi-rollout} (10$\times$, generated with 10 random seeds)~synthetic videos are generated for each data regime. These two sets are used to show if number of synthetic data matters. 

\noindent\textbf{IDM Training and Pseudo Action Labeling. } For IDM model training, we use the out-of-the-domain episodes together with varying 5, 10, 20~(thus three separate IDMs) training episodes. We started from the pretrained Franka IDM checkpoint from DreamGen~\cite{jang2025dreamgen} and finetuned the model for 10k steps with a learning rate of 1e-4. The IDM is then used to generate pseudo-labeled kinematics for the synthetic videos. \\
\noindent\textbf{Robot Policy Results.} For all the experiments, we start from the pretrained GR00T N1.5 checkpoint which is a strong starting point. Our base policy~(\textbf{Real Only}) is the model finetuned only with 5, 10, 20 real training data with learning rate 1e-4, 200 steps. For 56 and 560 synthetic data~(\textbf{Real+Synthetic}, \textbf{Real+Synthetic 10x}), we first finetune with learning rate 1e-4 for 400 steps, then we further finetune with 5, 10, 20 real training data with learning rate 1e-4 and 200 steps. We test these 6 policy models on the 40 held out test episodes. An example of left arm cartesian trajectory~(first 3 digits of the model's 20 dimension output) is shown in Fig.~\ref{fig:trajectory}. The trajectory (left arm movement) of Real + Syn 10x showed closer similarity to the groundtruth. We calculate the mean square error~(MSE) for all 20 dimension action prediction compared with groundtruth and averaged across all test data. The result is shown in Fig.~\ref{fig:mse_comparison}. We separate the cartesian, rotation and jaw since they have different physical meaning. We can see that the average MSE is the largest for the VLA finetuned only with real data, while with synthetic videos the average MSE gets lower. The trend holds for both varying real trajectory number and varying synthetic video number. Similar trend can be observed for varying data finetuning hyperparameters and for VLAs other then GR00T model~(e.g. $\pi_{0.5}$~\cite{intelligence2025pi}) , as shown in the supplementary.



\section{Conclusion}
\label{conclusion}
We present Cosmos-H-Surgical, the first surgical world model that connects high-quality synthetic surgical videos generation with robot action learning. By curating the SATA dataset and integrating inverse dynamics modeling, Cosmos-H-Surgical generates synthetic data that can improve downstream policy training, enabling scalable and safe surgical robot data curation. However, Cosmos-H-Surgical still has major limitations. It requires finetuning datasets from unseen robot embodiments for both world model and IDM, which require additional data curation efforts. Meanwhile, pseudo-kinematics from IDM still lack ground-truth precision and may introduce residual noise. Lastly, the current SATA dataset, though diverse and finely annotated for physical AI, is not covering all the publicly available datasets. Future work will focus on extending SATA with more complex and broader procedures and improving IDM for better pseudo-kinematic.

{
  \small
  \bibliographystyle{unsrt}
  \bibliography{main}
}
\clearpage
\section*{Supplementary Material}
\FloatBarrier
\setcounter{page}{1}


\begin{figure}
    \centering
    \includegraphics[width=1\linewidth]{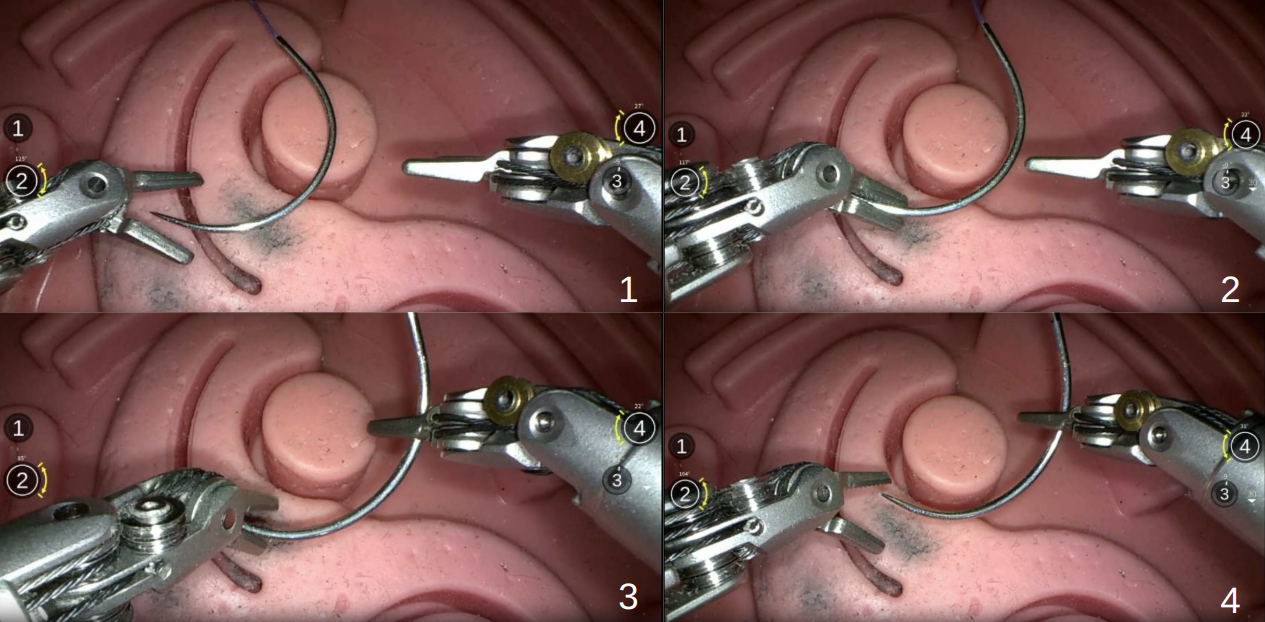}
    \caption{The illustration of the needle pick up and hand over task for the specific surgical robot platform.}
    \label{fig:needlepickup}
\end{figure}

\begin{figure*}[ht!]
    \centering
    \includegraphics[width=1\linewidth]{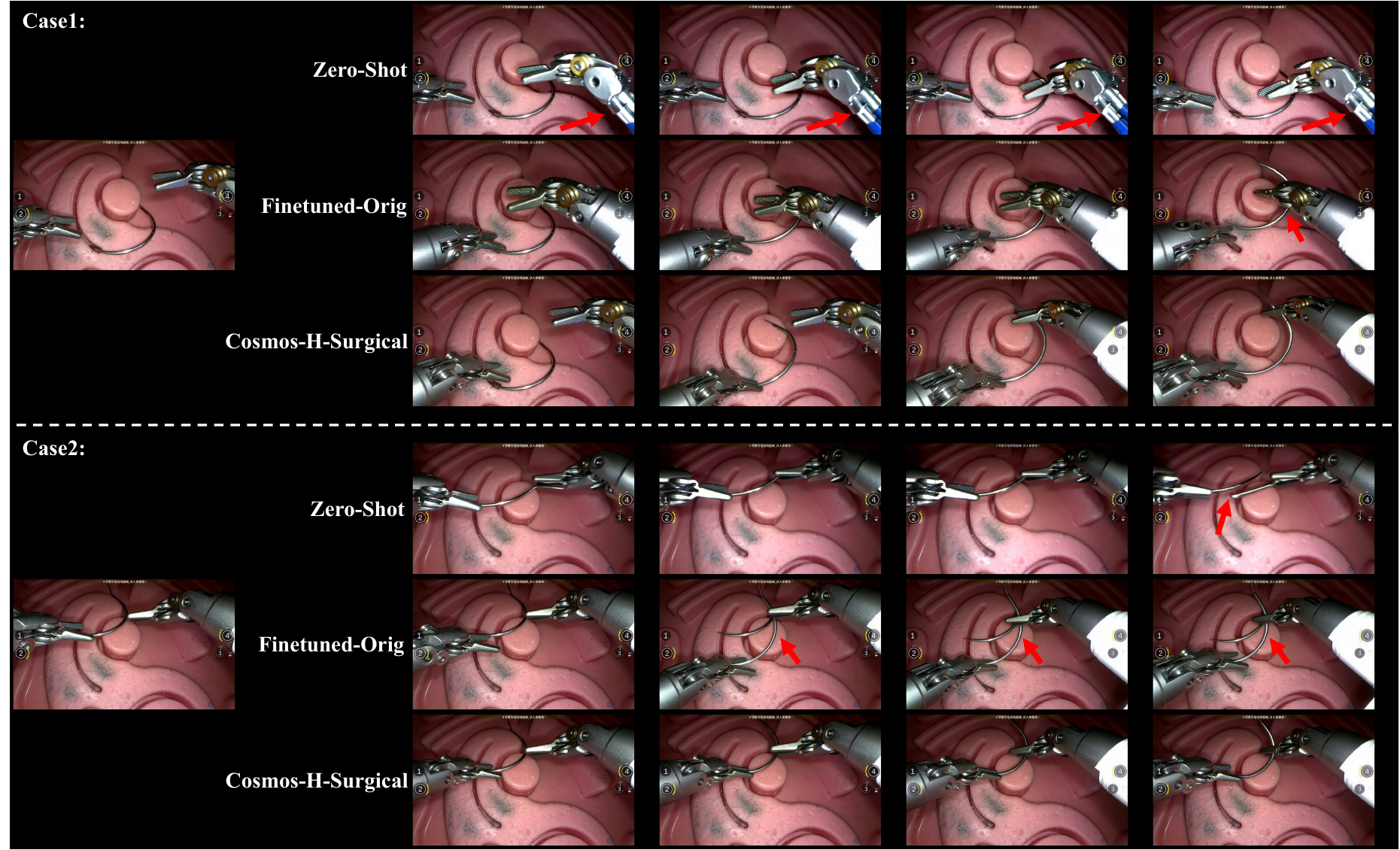}
    \caption{\textbf{Qualitative comparison of few-shot finetuning variants in real-world surgical scenes.}
We visualize predicted video rollouts from three model configurations—Zero-Shot, Finetuned-Orig, and Cosmos-H-Surgical—under two representative initial states (Case 1 and Case 2). Red arrows highlight incorrect surgical tools or hallucinations in the generated frames.}
    \label{fig:vis_few_shot}
\end{figure*}

\section{Human Expert Evaluation Criteria}
\label{sec:rationale}

\noindent\textbf{Human expert evaluation criteria.}
To assess perceptual realism and clinical fidelity, we conducted a human expert study with 3 surgical experts (one surgeon guides two residents to rate generated videos). Each expert independently evaluated 50 videos generated by different world model variants using a \textbf{three-level clinical quality rubric} (scores from 1 to 3, higher is better) across the following criteria:

\begin{itemize}[leftmargin=*]
    \item \textbf{Text–Video Alignment:} Assesses whether the generated scenes match the textual prompt and surgical viewing perspective. 
    \begin{itemize}
        \item \textbf{1:} Scene roughly aligns with the prompt but exhibits camera mismatch or unrealistic perspective.  
        \item \textbf{2:} Scene alignment is correct and transitions are mostly natural with minor motion jumps or deformations.  
        \item \textbf{3:} Scene fully aligns with prompt and surgical viewpoint; transitions are smooth and visually coherent.
    \end{itemize}

    \item \textbf{Tool Consistency:} Evaluates whether surgical instruments correspond to the prompt and remain physically consistent throughout the sequence.  
    \begin{itemize}
        \item \textbf{1:} Tool type matches the text but exhibits frequent deformation or discontinuity.  
        \item \textbf{2:} Tools remain largely consistent with reduced deformation; grasping and manipulation appear mostly realistic.  
        \item \textbf{3:} Tools behave continuously and naturally, accurately performing realistic grasping and needle handling without visual artifacts.
    \end{itemize}

    \item \textbf{Anatomical Structure:} Measures the plausibility of tissue and organ appearance and their reactions to tool interaction.  
    \begin{itemize}
        \item \textbf{1:} Basic anatomical layout correct but unrealistic structure or misaligned tissue response.  
        \item \textbf{2:} Improved detail reproduction with occasional deformation inconsistencies.  
        \item \textbf{3:} Anatomically accurate structures with realistic responses (e.g., traction, deformation, bleeding) closely resembling real surgery.
    \end{itemize}
\end{itemize}

\section{Cosmos-H-Surgical Training}
We adopt the Flow Matching (FM) formulation~\cite{lipman2022flow} to train the surgical video world model due to its conceptual simplicity and practical effectiveness. FM defines a velocity-based target in latent space, providing a direct training signal that improves optimization stability and sample quality. Formally, given a data sample $I$ (\eg, video frames), a noise vector $\epsilon \sim \mathcal{N}(0, I)$, and a timestep $t \in [0,1]$ sampled from a logit-normal distribution, the interpolated latent is defined as:
\begin{equation}
I_t = (1-t) I + t_\epsilon,
\end{equation}
where the corresponding ground-truth velocity is
\begin{equation}
v_t = \epsilon - I.
\end{equation}
The model predicts this velocity via a network $u_\theta(I_t, t, c)$, where $c$ represents the conditioning frame $I_0$ and corresponding text prompt, and parameters $\theta$ correspond to the trainable LoRA adapters. 
The flow-matching loss is then the mean squared error (MSE) between predicted and ground-truth velocities:
\begin{equation}
\mathcal{L}(\theta) = \mathbb{E}_{I, \epsilon, c, t} \big\| u_\theta(I_t, t, c) - v_t \big\|_2^2.
\end{equation}

\section{Additional Results}
\subsection{More Visual Comparisons}
To further illustrate the effect of few-shot adaptation on real surgical scenes, Figure~\ref{fig:vis_few_shot} presents qualitative comparisons across three model configurations, including Zero-Shot, Finetuned-Orig, and Cosmos-H-Surgical, evaluated under two representative initial conditions. Each rollout begins from a distinct endoscopic view with different needle and forceps arrangements, enabling a controlled examination of how well each model handles realistic surgical complexity.

The Zero-Shot model, which has no domain adaptation, frequently produces implausible motions and incorrect surgical tools, leading to early task failure. The Finetuned-Orig model reduces some hallucinations but still struggles with consistent grasping behavior and coherent tool interaction. In contrast, the Cosmos-H-Surgical model (SATA-pretrained then finetuned on five real trajectories) generates smooth and accurate motions that correctly execute the intended manipulation sequence. Red arrows mark common failure modes, including tool hallucination and incorrect action execution, emphasizing the improvement brought by domain-specific pretraining. These visual results underscore the importance of the proposed SATA-pretrained Cosmos-H-Surgical model for reliable few-shot generalization in real-world surgical environments.

\begin{table*}[t]
\centering
\small
\caption{\textbf{Source distribution of SATA video clips across four fundamental surgical actions.}}
\begin{tabular}{lcccc}
\hline
\textbf{Dataset} & \textbf{Knotting} & \textbf{Needle Grasping} & \textbf{Needle Puncture} & \textbf{Suture Pulling} \\ \hline
AutoLaparo~\cite{wang2022autolaparo}          & 47  & 9    & 42   & 30  \\
GraSP~\cite{ayobi2024pixel}               & 37  & --    & 1   & 28  \\
HeiCo~\cite{maier2021heidelberg}               & 16  & 3    & 6    & 4   \\
Multiypass140~\cite{ramesh2023weakly}      & 1   & --   & --   & --  \\
SAR-RARP50~\cite{psychogyios2023sar}           & 118 & 677  & 940  & 413 \\
SurgicalActions160~\cite{schoeffmann2018video}  & 7   & --   & --   & --  \\
YouTube~\cite{schmidgall2024general}             & 68  & --   & --   & --  \\ \hline
\end{tabular}
\label{tab:sata_source_distribution}
\end{table*}

\subsection{Additional Policy Results}
In this section we provide additional policy evaluation results. The task of ``Needle pick up and handover" is illustrated in Fig.~\ref{fig:needlepickup}.

\begin{figure*}
    \centering
    \includegraphics[width=1\linewidth]{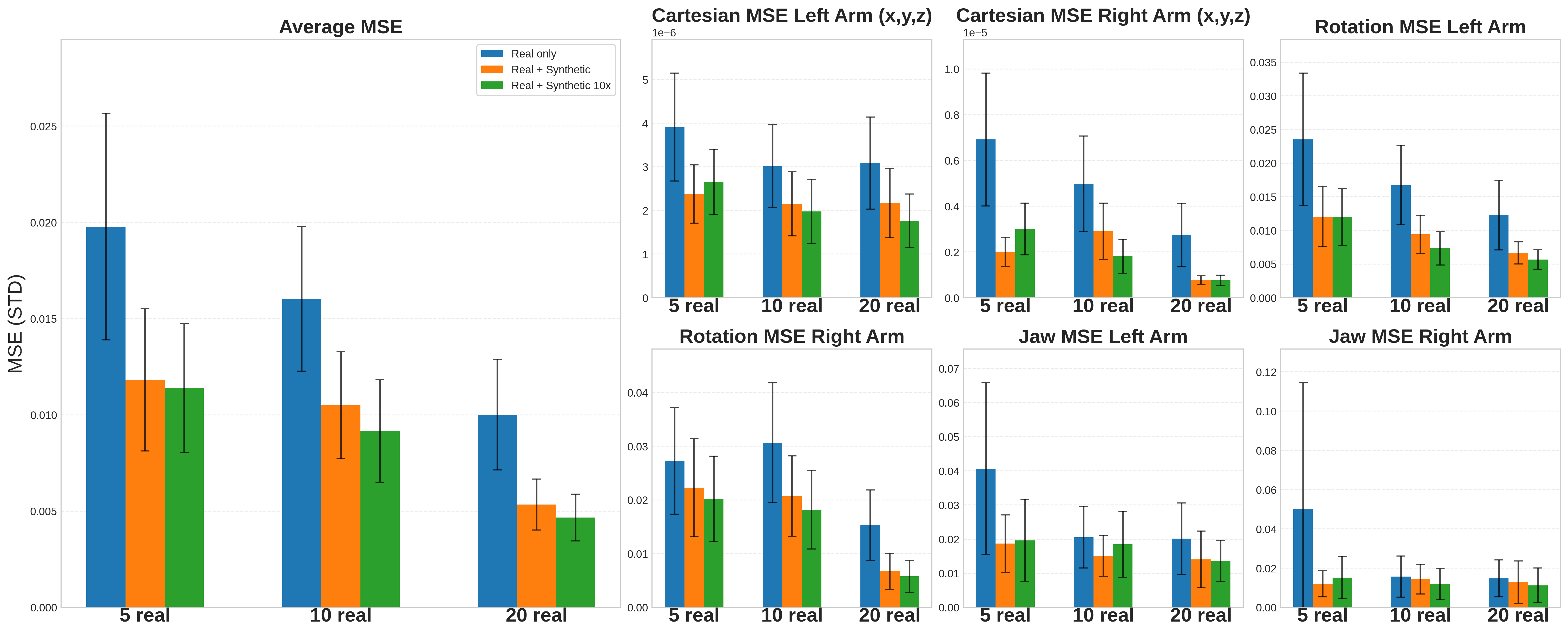} 
    \caption{\textbf{Multi-View real data finetuning}. Trajectory MSE (Standard deviation) on 40 test data. We use 5, 10, 20 real data with \textbf{3-View camera input} to finetune the policy, starting from GR00T N1.5 pretrained checkpoint (Real), and checkpoints pretrained from 56 (Real + Synthetic) and 560~(Real + Synthetic 10x) synthetic data~\textbf{Single View}.}
    \label{fig:mse_comparison_3view}
\end{figure*}

\noindent\textbf{Multi-view Robotic Policy} Internet video datasets typically contain only single-view endoscopic videos, and obtaining multi-view surgical videos for training a multi-view world model is currently not feasible. As a result, SurgeWorld generates only single-view synthetic videos. However, downstream surgical robots may use multiple cameras for multi-view policy learning. Here, we show that even when real surgical data includes multiple views—such as additional left and right wrist cameras—the single-view synthetic video–kinematic pairs can still improve multi-view VLA policy performance. We use the exact same training scheme and the same synthetic data as in the main paper, but we train on the 5, 10, and 20 real demonstrations while including the two additional wrist cameras, resulting in data with three views. Note that GR00T N1.5 VLA can process varying numbers of input views with the same weights. The results are shown in Fig.~.~\ref{fig:mse_comparison_3view}.

\begin{figure*}
    \centering
    \includegraphics[width=1\linewidth]{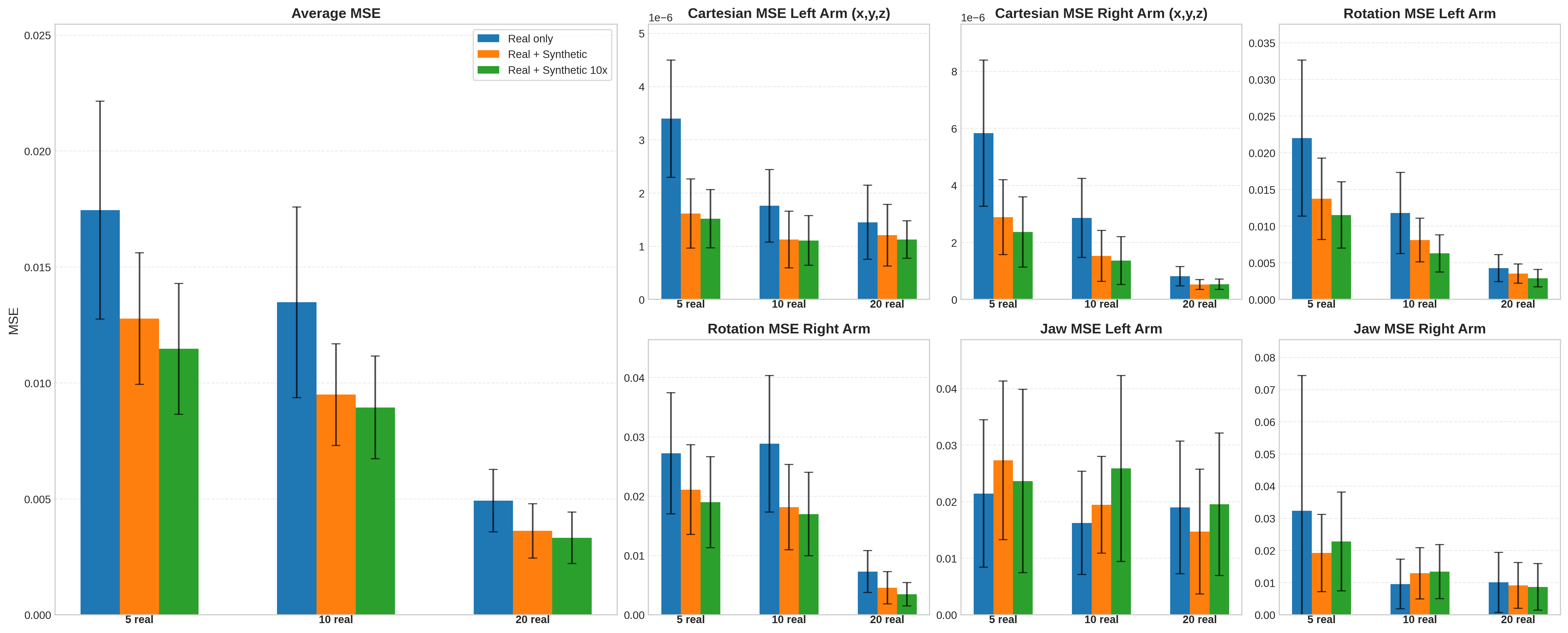} 
    \caption{\textbf{1k step finetuning}: Trajectory MSE (Standard deviation) on 40 test data. We use 5, 10, 20 real data to finetune the policy for \textbf{1k steps}, starting from GR00T N1.5 pretrained checkpoint (Real), and checkpoints pretrained from 56 (Real + Synthetic) and 560~(Real + Synthetic 10x) synthetic data.}
    \label{fig:mse_comparison_1k}
\end{figure*}

\begin{figure*}
    \centering
    \includegraphics[width=1\linewidth]{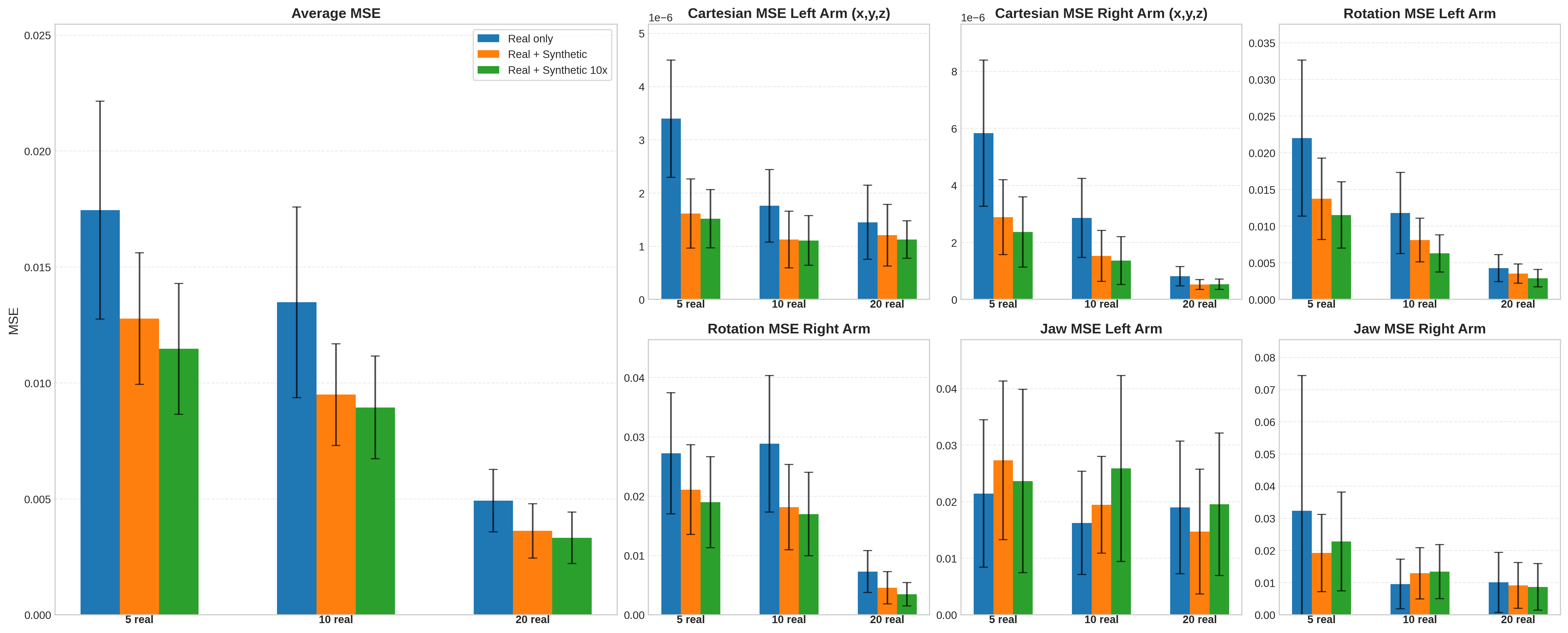} 
    \caption{\textbf{10k step finetuning}. Trajectory MSE (Standard deviation) on 40 test data. We use 5, 10, 20 real data to finetune the policy for \textbf{10k steps}, starting from GR00T N1.5 pretrained checkpoint (Real), and checkpoints pretrained from 56 (Real + Synthetic) and 560~(Real + Synthetic 10x) synthetic data.}
    \label{fig:mse_comparison_10k}
\end{figure*}

\noindent\textbf{Varying Hyperparameters} We also performed experiments varying the training steps. We repeated the GR00T policy training with varying 1k and 10k finetuning steps on the real data as shown in Fig.~\ref{fig:mse_comparison_1k} and Fig.~\ref{fig:mse_comparison_10k}. 

\begin{figure*}
    \centering
    \includegraphics[width=1\linewidth]{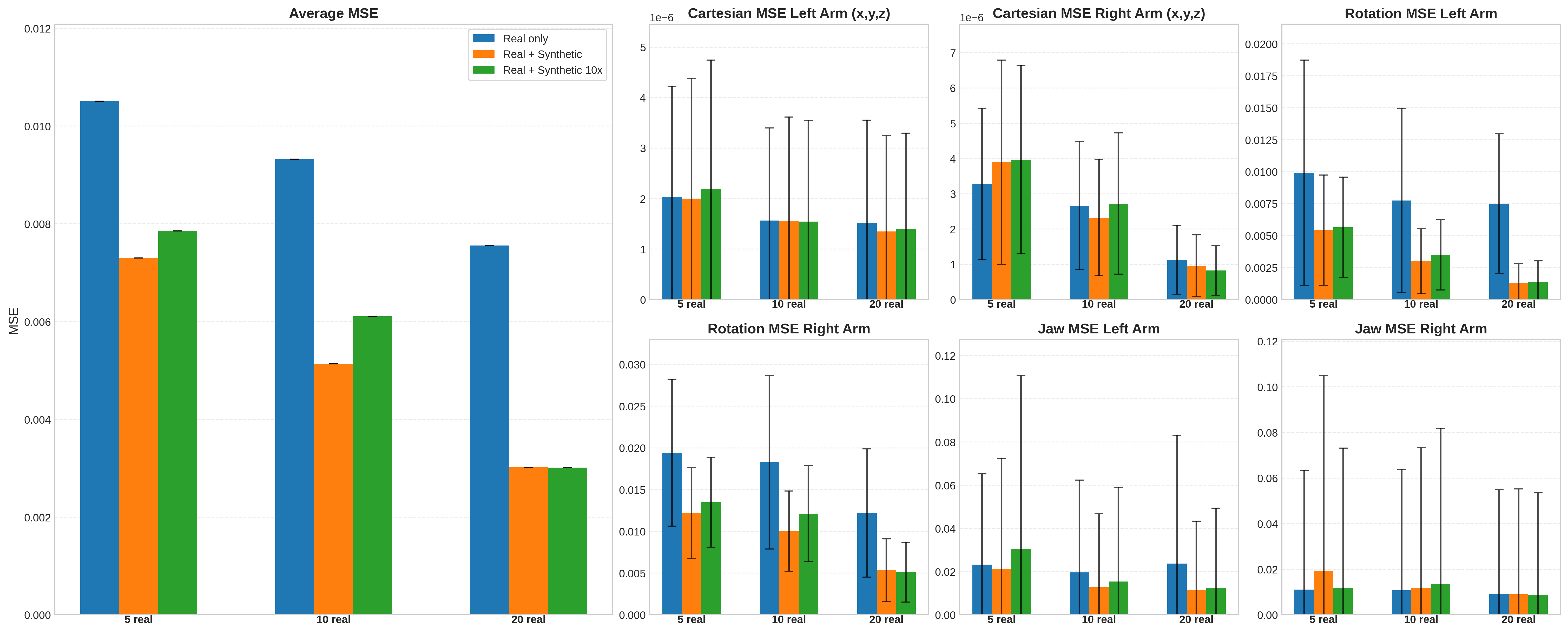} 
    \caption{$\boldsymbol{\pi_{0.5}}$ \textbf{results}. Trajectory MSE (Standard deviation) on 40 test data. We use 5, 10, 20 real data to finetune the policy, starting from GR00T N1.5 pretrained checkpoint (Real), and checkpoints pretrained from 56 (Real + Synthetic) and 560~(Real + Synthetic 10x) synthetic data.}
    \label{fig:mse_comparison_pi05}
\end{figure*}

\noindent\textbf{Other VLA} $\boldsymbol{\pi_{0.5}}$
To validate if Cosmos-H-Surgical and IDM can improve  other VLAs, we applied the same strategy to another recent foundational VLA model $\boldsymbol{\pi_{0.5}}$~\cite{intelligence2025pi}, which showed strong open world generalization. We simply change the GR00T policy to $\boldsymbol{\pi_{0.5}}$ and repeated the experiments as shown in  Fig.~\ref{fig:mse_comparison_pi05}.

\end{document}